# Leaf vein segmentation using Odd Gabor filters and morphological operations

Vini Katyal[1], Aviral[2]

[1] Computer Science, Amity University,
Noida, Uttar Pradesh, India
*vini_katyal@yahoo.co.in*

[2] Computer Science, Amity University,
Noida, Uttar Pradesh, India
*Kaviral12@yahoo.com*

**Abstract**
Leaf vein forms the basis of leaf characterization and classification. Different species have different leaf vein patterns. It is seen that leaf vein segmentation will help in maintaining a record of all the leaves according to their specific pattern of veins thus provide an effective way to retrieve and store information regarding various plant species in database as well as provide an effective means to characterize plants on the basis of leaf vein structure which is unique for every species. The algorithm proposes a new way of segmentation of leaf veins with the use of Odd Gabor filters and the use of morphological operations for producing a better output. The Odd Gabor filter gives an efficient output and is robust and scalable as compared with the existing techniques as it detects the fine fiber like veins present in leaves much more efficiently.

## 1. Introduction

According to American Society of Plant Physiologists the pattern and ontogeny of leaf venation appear to guide or limit many aspects of leaf cell differentiation and function. Photosynthetic, supportive, stomatal, and other specialized cell types differentiate in positions showing a spatial relationship to the vascular system. [1] The leaf vein patterns thus are a differentiating factor between various plant species and helps in analyzing their characteristics that would be suitable for medicinal purposes and maintaining a record of all the features that would be unique with respect to various plant species. The leaf patterns are unique for both Dicots and Monocots. Dicots essentially have a prominent midvein, several distinct vein size orders with smaller veins diverging from the larger, and a closed reticulum formed by several veins with the smallest veins forming so-called freely ending veinlets. The vein patterns of Dicots are essentially very diverse in comparison with monocots. Many approaches have been devised for leaf vein pattern recognition some of them are A Leaf Vein Extraction Method Based on Snakes Technique by Yun Feng Li Qing Sheng Zhu Yu Kun Cao Cheng Liang Wang. This methodology puts similar characteristics from the parametric and implicit deformable models in both the contour evolution process and the mechanisms for the contour guide [2], the Leaf vein extraction using Independent component analysis by Yan Li, Zheru Chi and David D. Feng where a FastICA is applied to leaf images to learn a set of linear basis functions or features for the images and then the basis functions are used as the pattern map for vein extraction [3]. Many methods also utilize the capability of edge detectors like sobel, pewitt and canny however the results are unsatisfactory in case of illuminations and noise. The algorithm that is proposed in the paper utilizes the Odd Gabor filters whose edge detection capability has been proven better as compared to other edge detectors [4]. This paper is organized as follows. The proposed automated leaf venation detection algorithm is given in Section 2 followed by the method used in Section 3. Finally, Experimental results and conclusion are given in Section 4 and Section 5 respectively.

## 2. The Proposed Algorithm

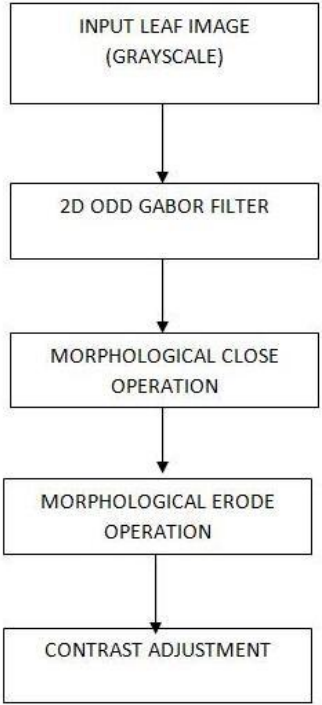

Figure1: PROPOSED ALGORITHM

## 3. Methods Used

### 3.1 Preprocessing

3.1.1 Collection of datasets
All the images were acquired from a 5 megapixel digital camera with uniform illumination conditions.

3.1.2 Pre-processing the leaf images
The images were converted to JPG format and then to grayscale format.

### 3.2 Odd Gabor filter

Gabor filters, which have been shown to fit well the receptive fields of the majority of simple cell in the primary visual cortex [5], are modulation products of Gaussian and complex sinusoidal signals. A 2D Gabor filter oriented at angle $\theta$ is given by:

$$g_{\theta,\vartheta,\varphi,\sigma,\gamma}(x,y) = \exp\left(-\frac{x^2 + \gamma^2 y^2}{2\sigma^2}\right)\cos\left(2\pi\frac{x^2}{\vartheta} + \varphi\right)$$

$$x' = x\cos\theta + y\sin\theta \qquad y' = -x\sin\theta + y\cos\theta \qquad (1)$$

$\lambda$: This is the wavelength of the cosine factor of the Gabor filter kernel and here with the preferred wavelength of this filter.

$\Theta$: This parameter specifies the orientation of the normal to the parallel stripes of a Gabor function.

$\Phi$: The phase offset $\varphi$ in the argument of the cosine factor of the Gabor function is specified in degrees. Valid values are real numbers between -180 and 180.

$\gamma$: This parameter, called more precisely the spatial aspect ratio, specifies the ellipticity of the support of the Gabor function.

$b$: The half-response spatial frequency bandwidth $b$ (in octaves) of a Gabor filter is related to the ratio $\sigma / \lambda$, where $\sigma$ and $\lambda$ are the standard deviation of the Gaussian factor of the Gabor function and the preferred wavelength, respectively, as follows:

$$b = \log_2 \frac{\frac{\sigma}{\lambda}\pi + \sqrt{\frac{\ln 2}{2}}}{\frac{\sigma}{\lambda}\pi - \sqrt{\frac{\ln 2}{2}}}, \quad \frac{\sigma}{\lambda} = \frac{1}{\pi}\sqrt{\frac{\ln 2}{2}} \cdot \frac{2^b + 1}{2^b - 1} \qquad (2)$$

The Odd Gabor filter is given by:

$$g_{\theta,\vartheta,\varphi,\sigma,\gamma}(x,y) = \exp\left(-\frac{x^2 + \gamma^2 y^2}{2\sigma^2}\right)\sin\left(2\pi\frac{x^2}{\vartheta} + \varphi\right)$$

$$x' = x\cos\theta + y\sin\theta \qquad y' = -x\sin\theta + y\cos\theta \qquad (3)$$

The Odd Gabor filter has been shown to be an efficient and robust edge detector [6] which offers distinct advantages over traditional edge detectors, such as Roberts, Sobel, etc., and can be comparable even superior to Canny edge detector [7] generally thought as an optimal edge detector. Thus the algorithm utilizes the capabilities of the Odd Gabor filters to effectively detect the leaf veins.

| S.No | Parameters | Values used |
|---|---|---|
| 1. | Wavelength($\lambda$) | 8 |
| 2. | Orientation(s) ($\theta$) | 0, 30, 45, 60, 90, 120 180 |
| 3. | Phase offset(s) ($\varphi$) | 0 |
| 4. | Aspect ratio ($\gamma$) | |
| 5. | Bandwidth ($b$) | 5 |
| 6. | Number of orientations | 8 |

Table 1: Shows the optimal parameters for leaf vein detection

### 3.3 Morphological Close operation

Closing is an important operator from the field of morphology. The effect of this operator is to preserve background regions that have a similar shape to the structuring element, while eliminating all the regions of the background pixels it can be considered equivalent to dilation but it causes less destruction. The closing operator therefore requires two inputs: an image to be closed and a structuring element. Gray level closing consists straightforwardly of a gray level dilation followed by gray level erosion. Closing is the dual of opening, *i.e.* closing the foreground pixels with a particular structuring element, is equivalent to closing the background with the same element [8].

The Close operation is given by the following equation:

$$A \bullet S = (A \oplus S) \ominus S \qquad (4)$$

where A is the input image and S the structuring element.

The structuring element that has been utilized in the algorithm is the disk element

Se= strel('disk',3);

Im =imclose(img_out_disp, se);

The Close operation is utilized to fill in the gaps between the double edges which arise with Odd Gabor filters which is an un desired feature with edge detectors. The Leaf veins are effectively converted into single edges with the close operation.

The close operation is very effective in filling the gaps and fast in execution.

### 3.4 Morphological Erosion

Erosion is one of the basic operators in the area of mathematical morphology. The basic effect of the operator on a binary image is to erode away the boundaries of regions of foreground pixels. Thus areas of foreground pixels shrink in size, and holes within those areas become larger [9]. The basic erosion is given by:

$$A \ominus S = \{x : (B)_x \subseteq A\} = \bigcap_{x \in S} A_x \qquad (5)$$

A is the input image and S the structuring element. This is also called the Minkwoski Subtraction.

Erosion has been used to get rid of the unwanted pixels present next to the veins and highlight the leaf veins. Thresholding followed by skeletonizing and thinning has not been used as it removes the important leaf veins and the output retrieved isn't as efficient.

A structuring element of Line has been used taking in view that the leaf veins are directionally oriented lines.

se1=strel('line',2,0);
im1=imerode(im,se1);

### 3.5 Contrast adjustment

Contrast adjustment maps the intensity values in grayscale input image to new values in the image such that 1% of data is saturated at low and high intensities of the input image. This increases the contrast of the output image. The results got after Morphological erosion doesn't have a good contrast i.e. there are no sharp differences between black and white pixels (MATLAB help)

im2=imadjust(im1);

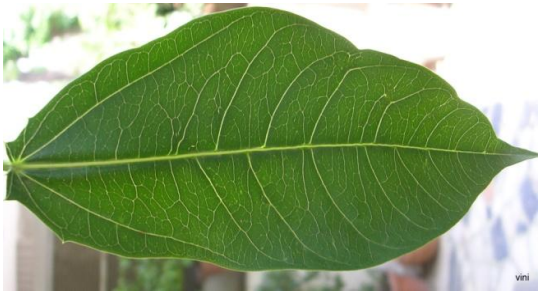

(a)

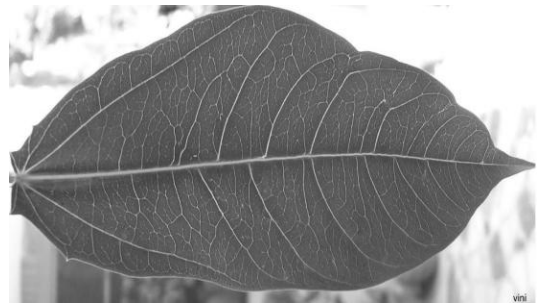

(b)

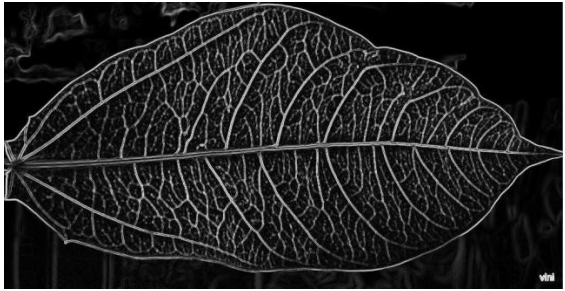 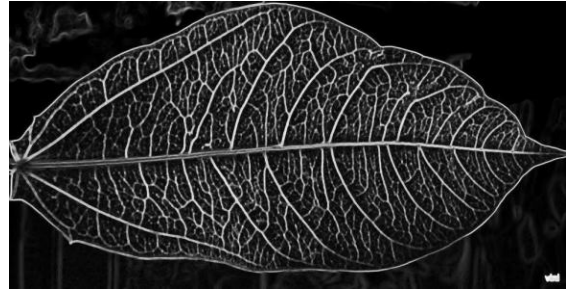

(c) (d)

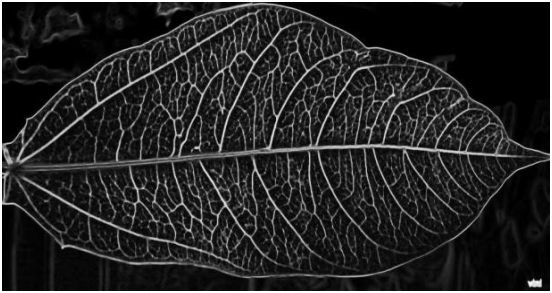 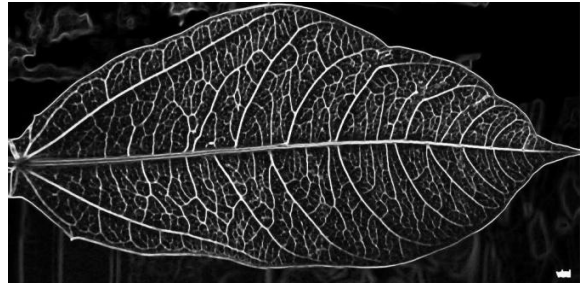

(e) (f)

Figure 2: Gives the Leaf vein extraction using the algorithm

## 4. EXPERIMENTAL RESULTS

The experiments have been carried out using various leaves and the outputs are extremely accurate. The algorithm is extremely fast and robust its evaluation time is found to be around 12 seconds for large images and less than 10 seconds for small images.

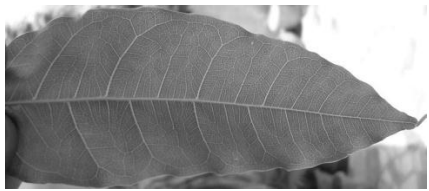 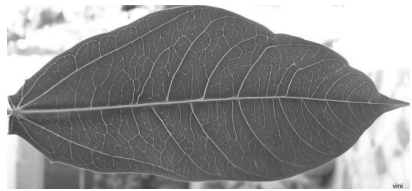 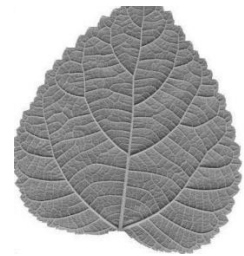

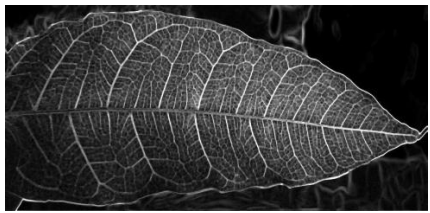 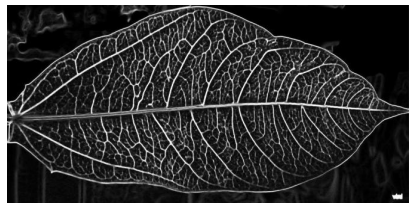 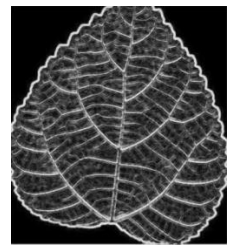

Figure 3: Gives the results on various leaves by the use of the algorithm

## 5. CONCLUSION

Leaf venation is important for identifying plants. These structures are always species specific and will consistently grow to a genetically determined pattern and shape. Botanists and foresters have developed terms for these patterns and shapes which help in tree identification. Thus the identification and segmentation is of utmost importance which is presented by the approach in the algorithm. This method is flexible as compared to other approaches proposed so far.

## 6. REFERENCES

[1] Timothy NelsonaY1, Nancy Denglerb," Leaf Vascular Pattern Formation", The Plant Cell, Vol. 9, 1121-1 135, July 1997 O 1997 American Society of Plant Physiologists.

[2] Yun Feng Li Qing Sheng Zhu Yu Kun Cao Cheng Liang Wang, "A Leaf Vein Extraction Method Based On Snakes Technique", IEEE 2005.

[3] Yan Li, Zheru Chi, "Leaf Vein Extraction Using Independent Component Analysis", 2006 IEEE International Conference on Systems, Man, and Cybernetics October 8-11, 2006, Taipei, Taiwan.

[4] Zhenfeng Zhu, Hanqing Lu, Yao Zhao, "Scale multiplication in odd Gabor transform domain for edge detection", Journal of Visual communication and image representation, Volume 18 Issue 1, February, 2007

[5] J.G. Daugman, Complete discrete 2-D Gabor transforms by neural networks for image analysis and compression, IEEE Trans. Acoust. Speech Signal Processing. 36 (7) (1988) 1169–1179.

[6] R. Mehrotra, K.R. Namuduri, N. Ranganathan, "Odd Gabor filter-based edge detection", Pattern Recognit. 25 (12) (1992) 1479–1494.

[7] J.F. Canny,"A computational approach to edge detection", IEEE.Trans. Pattern Anal. Mach. Intell. 8 (6) (1986) 679–698.

[8] R. Gonzalez and R.Woods, Digital Image Processing,Addison-Wesley Publishing Company, 1992, pp 524, 552.

[9] R. Haralick and L. Shapiro, Computer and Robot Vision, Vol. 1, Chap. 5, Addison-Wesley Publishing Company, 1992.

[10] A. Jain, Fundamentals of Digital Image Processing, Prentice-Hall, 1986, p 384.

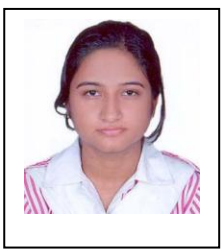

Vini Katyal

Currently pursuing her Bachelor's in Computer Science and technology from Amity University, Uttar Pradesh and keeps interest in Image processing.

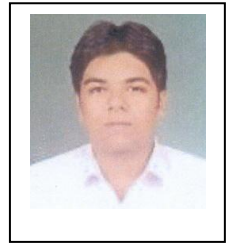

Aviral

Currently pursuing his Bachelor's in Computer Science and technology from Amity University, Uttar Pradesh and has interest in Computer vision